\documentclass[lettersize,journal]{IEEEtran}
\usepackage{array}

\usepackage{makecell}
\usepackage{booktabs}
\usepackage{textcomp}
\usepackage{stfloats}
\usepackage{url}
\usepackage{verbatim}
\usepackage{cite,soul}
\usepackage{amsmath,amssymb,amsfonts}
\usepackage{algorithmic}
\usepackage{graphicx}
\usepackage{subfigure}
\usepackage{xcolor}
\usepackage[ruled]{algorithm2e}
\usepackage{booktabs}
\usepackage{multirow}
\usepackage{epstopdf}
\usepackage{hyperref}
\usepackage{graphicx}
\usepackage{amsmath, amsthm, amsfonts}
\usepackage{amsmath}

\usepackage{amssymb}
\usepackage{siunitx}
\usepackage{etoolbox}

\makeatletter 
\pretocmd\@bibitem{\color{black}\csname keycolor#1\endcsname}{}{\fail}
\newcommand\citecolor[1]{\@namedef{keycolor#1}{\color{blue}}}
\makeatother

\begin{document}
\title{SocialTraj: Two-Stage Socially-Aware Trajectory Prediction for Autonomous Driving \\
via Conditional Diffusion Model}


\author{Xiao Zhou, Zengqi Peng, and Jun Ma, \textit{Senior Member, IEEE}
\thanks{Xiao Zhou and Zengqi Peng are with the Robotics and Autonomous Systems Thrust, The Hong Kong University of Science and Technology (Guangzhou), Guangzhou, China. (e-mail: xzhou910@connect.hkust-gz.edu.cn; zpeng940@connect.hkust-gz.edu.cn) }
\thanks{Jun Ma is with the Robotics and Autonomous Systems Thrust, The Hong Kong University of Science and Technology (Guangzhou), Guangzhou, China, and also with the Division of Emerging Interdisciplinary Areas, The Hong Kong University of Science and Technology, Hong Kong SAR, China (e-mail: jun.ma@ust.hk).} 
}



\maketitle

\begin{abstract}
Accurate trajectory prediction of surrounding vehicles (SVs) is crucial for autonomous driving systems to avoid misguided decisions and potential accidents. However, achieving reliable predictions in highly dynamic and complex traffic scenarios remains a significant challenge. One of the key impediments lies in the limited effectiveness of current approaches to capture the multi-modal behaviors of drivers, {which leads} to predicted trajectories that deviate from actual future motions.  To address this issue, we propose SocialTraj, a novel trajectory prediction framework integrating social psychology principles through social value orientation (SVO). 
By utilizing Bayesian inverse reinforcement learning (IRL) to estimate the SVO of SVs, we obtain the critical social context to infer the future interaction trend. 
To ensure modal consistency in predicted behaviors, the estimated SVOs of SVs are embedded into a conditional denoising diffusion model that aligns generated trajectories with historical driving styles.
Additionally, the planned future trajectory of the ego vehicle (EV) is explicitly incorporated to enhance interaction modeling. Extensive experiments on NGSIM and HighD datasets demonstrate that SocialTraj is capable of adapting to highly dynamic and interactive scenarios while generating socially compliant and behaviorally consistent trajectory predictions, outperforming existing baselines. Ablation studies {demonstrate} that dynamic SVO estimation and explicit ego-planning components notably improve prediction accuracy
and substantially reduce inference time.

\end{abstract}


\section{Introduction}

\IEEEPARstart{I}{n} recent years, autonomous driving technology has witnessed remarkable advancements, with intelligent systems becoming increasingly proficient in perception, planning, and control \cite{10401938,10329446,10285467,wang2024chance}. Among these capabilities, trajectory prediction plays a pivotal role in ensuring safe and efficient driving behaviors by anticipating the future behaviors of surrounding vehicles (SVs) \cite{10269660,10356826}. However, accurate trajectory prediction in highly dynamic and complex traffic scenarios remains a significant challenge.
{The dynamic nature of driving arises from continuous interactions between vehicles, as the movement of one vehicle alters the responses of others. In addition, variations in driver intent, risk tolerance, and social preferences produce heterogeneous driving styles that result in distinct maneuvers under similar conditions. \cite{peng2024reward,peng2025bilevel} }
This behavioral diversity introduces significant multi-modality in future trajectories, complicating the trajectory prediction task for the ego vehicle (EV).

\IEEEpubidadjcol
\begin{figure*}[!t]
\centering{\includegraphics[trim={0 0 0 1cm},width=\textwidth]{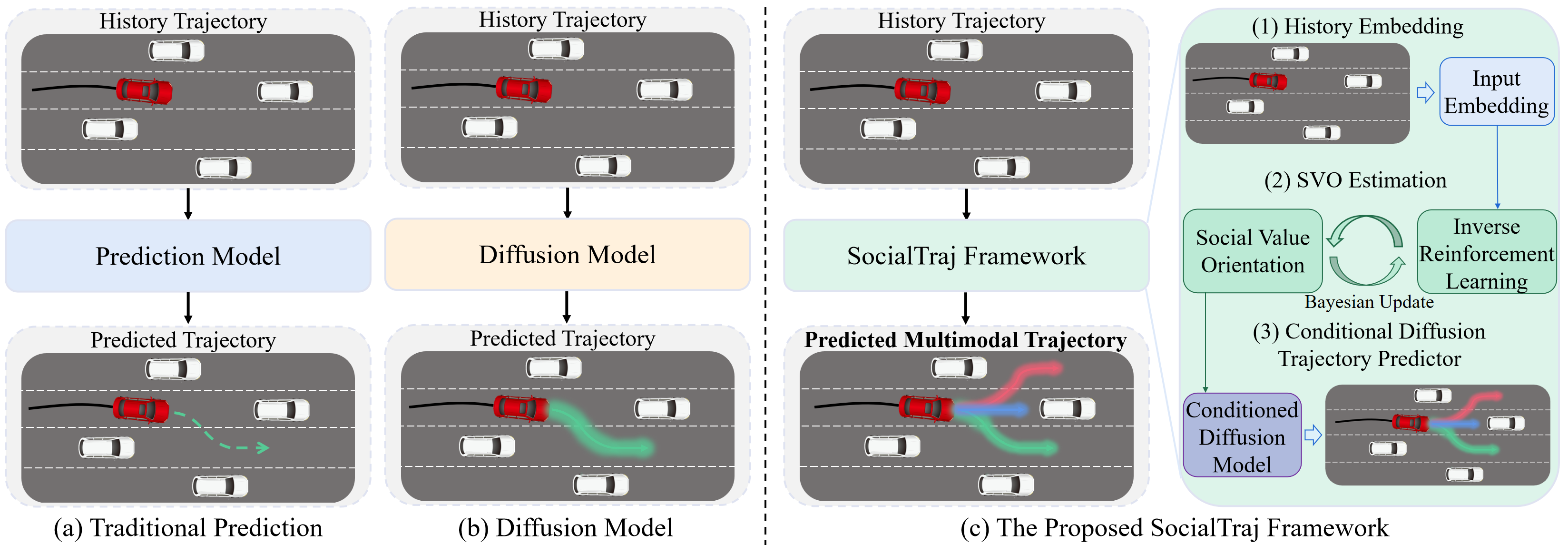}} 
\caption{Overview of existing and our proposed trajectory prediction frameworks. In (a), traditional methods directly map past trajectories to a single future trajectory through supervised learning, without iterative refinement. In (b), diffusion‐based approaches gradually generate future trajectories through the denoising process to obtain a prediction distribution. In (c), our SocialTraj framework first embeds history trajectories, then estimates the SVO of each agent via Bayesian IRL, and finally conditions a diffusion process on the inferred SVO and future planned trajectory of the EV to generate diverse and socially coherent future trajectories.}
\label{Overview}
\end{figure*}

{Traditional trajectory prediction methods typically treat each vehicle's trajectory independently, extrapolating future motion from historical kinematic data. Early approaches utilized recurrent neural networks (RNNs) with long short-term memory (LSTM) units to encode temporal dependencies from past positions and velocities \cite{altche2017lstm}. 
Convolutional encoder–decoder architectures similarly {capture} spatio-temporal features by applying sliding-window filters over trajectory segments \cite{deo2018convolutional}. 
Although effective in sparse scenarios, these methods often fail in dense environments due to their inability to model interactive maneuvers such as evasive actions or sudden braking induced by SVs. Graph neural networks (GNNs) {are} subsequently introduced to explicitly encode spatial interactions between agents \cite{singh2022graph}, complemented by attention mechanisms that dynamically weigh neighbor and lane features \cite{messaoud2020attention,lin2021vehicle,zhou2024integrated}. More recently, Transformer-based predictors have further advanced interaction modeling through self-attention layers that capture long-range dependencies \cite{geng2023physics,chen2022vehicle}. Nevertheless, due to reliance on pointwise regression losses, these methods frequently produce mean-centered forecasts, failing to capture the inherently multi-modal trajectory distributions, particularly in dense traffic conditions where potential trajectories diverge significantly \cite{rhinehart2018deep}.}

Complementing these traditional approaches, generative models have been proposed to capture the inherent multi-modality of human driving. 
Mixture density networks (MDNs) frame trajectory prediction as learning a mixture distribution over future states, directly modeling uncertainty in a probabilistic output layer \cite{chen2020predicting}. Generative adversarial networks (GANs) adopt an adversarial objective to encourage diverse and realistic samples \cite{rossi2021vehicle,hegde2020vehicle}, whereas conditional variational autoencoders (CVAEs) leverage latent variables to disentangle modality in trajectory space \cite{ivanovic2020multimodal}. 
{Inspired by recent breakthroughs in diffusion probabilistic models, researchers have begun exploring denoising diffusion approaches for trajectory generation \cite{croitoru2023diffusion} and prediction \cite{10352973}. Diffusion models offer a principled mechanism for sampling from complex, multi-modal distributions, iteratively refining noisy proposals into coherent trajectories that mitigate mode collapse and blur shown in GANs and CVAEs. 
{In \cite{wang2024optimizing}, optimal Gaussian diffusion (OGD) and estimated clean manifold guidance are proposed to accelerate the reverse process to practical runtimes}, yet this approach still treats all agents symmetrically and omits explicit modeling of ego‐plan effects. 
The DICE framework proposed in \cite{10588780} combines a scoring network with a diffusion backbone to rank high‐probability pedestrian and vehicle trajectories. 
But its universal scoring tends to over-penalize cooperative maneuvers in dense traffic, leading to conservative predictions.  
Equivariant diffusion networks (EquiDiff) embed SO(2) symmetry into the denoiser to improve geometric consistency with the denoising diffusion probabilistic model (DDPM)  \cite{chen2023equidiff}, yielding robust short‐term forecasts but suffering from degraded long‐term accuracy when social context shifts rapidly.  More recently, Crossfusor \cite{you2024crossfusor} employs cross-attention transformers within conditional diffusion for car following, capturing local inter-vehicle dependencies. 
Although those works effectively capture geometric consistency and local spatial dependencies, their reliance on fixed or heuristically selected conditioning features limits their capability to model dynamic vehicle interactions, which is an essential factor for realistic autonomous driving scenarios.
}

{
{To address this limitation, interactive trajectory prediction is investigated to explicitly model the vehicle interactions, including classic, game-theoretic, and data-driven methods.} 
Classical social-force methods represent each vehicle as a particle influenced by attractive and repulsive forces to maintain gaps and model car-following behaviors \cite{treiber2000congested,helbing1995social}. Extensions have employed Lagrangian mechanics or neural ordinary differential equations to infer driver intent from smooth potential fields or continuous dynamics \cite{xie2013inferring,liang2021modeling}. While these rule-based models offer interpretability, their fixed interaction kernels and uniform sensitivity parameters hinder generalization to the heterogeneous, context-dependent maneuvers seen in real traffic.
To preserve interpretability while handling the richer interactions of dense real-world scenarios, the researchers {adopt} game-theoretic frameworks. 
In the Stackelberg formulation, the leader vehicle plans its trajectory under the assumption that followers will respond optimally \cite{schwarting2019social}. Level-$k$ reasoning method models agents who iteratively compute their best responses within a bounded rationality \cite{10700678}.
These approaches naturally capture mutual foresight and phenomena such as yielding and blocking. However, game-theoretic approaches become computationally intractable in continuous, high-dimensional settings and are highly sensitive to incorrect assumptions about the decision-making model of other drivers.
Data-driven methods address both the rigidity of rule-based kernels and the complexity of strategic solvers by learning social dependencies directly from trajectories. Social-LSTM and its variants pool neighbors’ hidden states to implicitly encode interactions \cite{alahi2016social,gupta2018social}, while GNNs model vehicles as nodes with learned edge weights \cite{xu2022adaptive}. These techniques scale efficiently and achieve high predictive accuracy, but they encapsulate social information in latent spaces, sacrificing interpretability and lacking mechanisms to adjust influences as driver intent evolves.
To {integrate} interpretability with real-time adaptability, we leverage social value orientation (SVO), a well-validated social-psychology construct that quantifies the balance of an individual between self-interest and collective welfare. 
However, previous methods with the SVO model \cite{peng2021learning,10634799} cannot capture the rapid changes in cooperative or competitive driving behaviors in real-time.
}

With the above descriptions as a backdrop, we present a socially aware and interactive {two-stage} trajectory prediction framework for autonomous driving (as shown in Fig. \ref{Overview}).
Specifically, the main contributions  of the paper are as follows:
{
\begin{itemize}
    \item 
    {We propose SocialTraj, a trajectory prediction framework that leverages an attention-based Bayesian inverse reinforcement learning (IRL) approach to infer the SVO of SVs in real time,}  enabling accurate and dynamic updates of SVO estimations.
    \item By integrating the estimated SVO and the planned future trajectory of the EV into a conditional DDPM, we explicitly capture inter-vehicle interactions and generate socially coherent, behaviorally aligned trajectory predictions.

    \item Extensive experiments on the NGSIM and HighD datasets demonstrate SocialTraj’s superior performance across all metrics. Ablation studies further validate that dynamic SVO estimation and explicit ego-planning enhance prediction accuracy
    and significantly reduce inference time.
\end{itemize}
}

The remainder of this paper is organized as follows: Section II covers the necessary preliminaries, including our social interaction model, IRL, and DDPM. Our methodology is formally presented in Section III, detailing the Bayesian IRL approach for SVO estimation, describing the conditional diffusion model for trajectory prediction, and explaining our joint training pipeline. As outlined in Section IV, we report and analyze experimental results on two real-world datasets and further present ablation studies of core components in our method. Finally, we conclude the paper in Section V and discuss future directions.


\section{Preliminaries}

\subsection{Social Interaction Model}
\begin{figure}[]
  \centering
  \includegraphics[width=\linewidth]{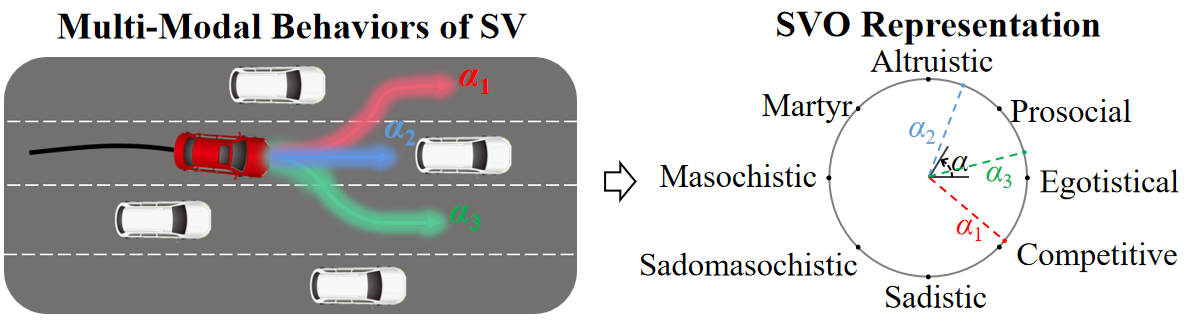}
  \caption{Illustration of SVO for modeling multi-modal driving behaviors.  
  The three SVO values  \(\alpha_1,\alpha_2,\alpha_3\) represent different possible trajectory modalities of SV.
  Mapping each trajectory mode to an SVO angle provides a compact, psychologically grounded representation of driver intent that guides socially coherent prediction.}
  \label{fig:svo_representation}
\end{figure}

SVO originates from social psychology as a metric of how individuals trade off their own benefit against that of others in shared decision-making contexts \cite{murphy2011measuring}.
Geometrically, the SVO of each agent can be represented by an angle \( \alpha \in [0, 2 \pi) \) in a two–dimensional payoff space as shown in Fig. \ref{fig:svo_representation}, where the horizontal axis corresponds to the reward of the ego agent and the vertical axis to the aggregate reward of surrounding agents. By projecting a resource‐allocation vector onto these axes, $\alpha$ succinctly encodes the propensity of an agent toward self‐interest or cooperation.

Within this framework, three archetypal orientations emerge. 
When $\alpha$ lies near $45^\circ$ , the agent is prosocial, so it balances self interest with group benefit,  with a behavior characteristic of courteous merging or yielding. Conversely, an individualistic orientation (\(\alpha \approx 0^\circ\)) places maximal weight on personal payoff, often manifesting as aggressive lane‐keeping. 
At the other extreme, competitive agents (\( \alpha > 270^\circ \)) privilege the reduction of others’ rewards, leading to blocking or adversarial maneuvers.

In this work, we parameterize the SVO of each vehicle by an angle $\alpha_i$  that reflects the ratio between expected individual reward and  global reward as:
\begin{equation} \label{eq:svo}
\alpha_i = \arctan\left( \frac{\mathbb{E}[R_{\text{g}}^i]}{\mathbb{E}[R_{\text{i}}^i]} \right)
\end{equation}
where $R_{\text{g}}^i$ and $ R_{\text{i}}^i$ represent the individual rewards and global rewards for agent $i$ and the expectations are derived from historical interaction patterns.
During traffic interactions, agents select trajectories by maximizing their reward.
Therefore, {the trajectories of vehicle $i$} are governed by a reward function:  
\begin{equation}
\label{reward_func}
R^i_{\alpha} = R_{\text{i}}^i \cdot \cos\alpha + R_{\text{g}}^i \cdot \sin\alpha
\end{equation}

In this work, we assume SVO remains consistent during short-term interactions. This assumption aligns with empirical observations in crowded navigation scenarios, where the social preferences of agents exhibit temporal consistency over limited horizons \cite{crosato2022interaction}. By integrating SVO into trajectory prediction, our framework aims to capture the multi-modal behavior of human drivers, bridging the gap between geometric trajectory planning and socio-psychological decision models.

\subsection{Inverse Reinforcement Learning}
IRL {serves} to infer the hidden objectives that govern observed driving behaviors, framing each vehicle as an optimal planner acting under an unknown reward function rather than prescribing driver preferences by hand. In our setting, we posit that an exhibited trajectory of SV encodes its implicit trade-offs among safety, efficiency, and comfort, which we aim to recover through IRL to estimate its underlying SVO.

To operationalize this, we model the decision process of each vehicle as a Markov Decision Process (MDP) defined by the tuple
\(
\mathcal{G}_{\mathrm{MDP}} = \langle \mathcal{S}, \mathcal{A}, \mathcal{P}, R_\alpha, \gamma \rangle,
\)
where \(\mathcal{S}\) denotes the state space of vehicle motion feature, \(\mathcal{A}\) is the action space of control inputs, \(\mathcal{P}(s' \mid s,a)\) signifies the transition kernel, \(\gamma\in(0,1)\) stands for the discount factor, and \(R_\alpha : \mathcal{S}\times\mathcal{A}\to\mathbb{R}\) refers to the reward function. Specifically, we adopt the reward function defined in (\ref{reward_func}), 
thereby, the SVO angle $\alpha\in [0, 2 \pi)$ denotes the reward parameter vector.
Our goal is to recover the reward parameter \(\alpha^*\) that best explains a given expert trajectory \(\tau = (s_0, a_0, s_1, a_1, \dots, s_{T-1}, a_{T-1})\), where $T$ is the trajectory
length. Under the maximum‐entropy IRL framework, the probability of \(\tau\) is defined as:
\begin{equation}
P_\alpha(\tau\mid\alpha)
= \frac{1}{Z(\alpha)} \exp\!\biggl(\sum_{t=0}^{T-1} R_\alpha(s_t, a_t)\biggr)
\end{equation}
{where
$Z(\alpha) = \sum_{\tau'\in\mathcal{T}}
    \exp\!\Bigl(\sum_{t=0}^{T-1}R_\alpha(s'_t,a'_t)\Bigr)$
is the partition function that normalizes over all possible trajectories $\mathcal{T}$. We then estimate \(\alpha^*\) by maximizing the regularized log‐likelihood of the expert demonstration as:
\begin{equation}
\alpha^* 
= \arg\max_{\alpha}\; \Bigl[\log P_\alpha(\tau\mid\alpha) \;-\; \lambda \|\alpha\|^2\Bigr]
\end{equation}
where $\lambda$ is a regularization weight. Gradient‐based optimization of this concave objective yields \(\alpha^*\), thereby recovering a reward model that explains observed driving behaviors.}

Although IRL theoretically enables interpretable modeling of driver intent, its application to real‐world traffic requires addressing high‐dimensional continuous state–action spaces and non‐stationary interaction patterns. To this end, we adopt a Bayesian IRL extension that maintains a posterior distribution over \(\alpha\), allowing online updates as new trajectory data become available. This belief‐driven inference not only captures uncertainty in SVO estimations but also facilitates adaptive refinement of predicted behaviors in dynamic traffic environments.

\subsection{Denoising Diffusion Probabilistic Model}

DDPMs formulate generation as the reversal of a gradual noising process, thereby learning an expressive approximation to complex data distributions \cite{ho2020denoising}. In the \emph{forward diffusion}, a clean sample \(x_0\) is progressively corrupted into Gaussian noise through a predefined cumulative noise schedule \(\{\bar\phi_t\}_{t=1}^{T_s}\), 
{where ${T_s}$ is number of diffusion steps}. 
Concretely, at each timestep \(t\) the data is perturbed according to
\begin{equation}
x_t \;=\; \sqrt{\bar\phi_t}\,x_0 \;+\; \sqrt{1-\bar\phi_t}\,\epsilon,
\quad \epsilon\sim\mathcal{N}(0,I)
\end{equation}
so that as \(t\to {T_s}\), \(x_{T_s}\) approaches an isotropic Gaussian. Modern implementations favor cosine‐based schedules to allocate noise more evenly across timesteps, improving stability in early denoising stages \cite{nichol2021improved}.

Reconstruction proceeds by training a neural network \(\epsilon_\psi(x_t,t)\) to predict the added noise at each level of corruption. {Here, $\psi$ represents the set of conditional features including historical observations, estimated SVO, and future planned trajectory of the EV}. Equivalently, the model learns the conditional denoising kernels
\begin{equation}
p_\psi(x_{t-1}\mid x_t)
=\mathcal{N}\bigl(\mu_\psi(x_t,t),\,\Sigma_t\bigr)
\end{equation}
where \(\Sigma_t\) may be fixed from the forward process. Optimization of the simplified objective
$\mathcal{L}
=\mathbb{E}_{t,\,x_0,\,\epsilon}
\bigl[\|\epsilon-\epsilon_\psi(x_t,t)\|^2\bigr]$.
yields a stable estimator of the noise residual, which in turn allows exact recovery of \(x_0\) via iterative denoising.

At inference time, new samples are generated by initializing \(x_{T_s}\sim\mathcal{N}(0,I)\) and applying ${T_s}$ reverse‐diffusion steps. Each step subtracts the noise estimation of the network and rescales the result, gradually transforming pure noise into a data‐like sample. In our SocialTraj framework, conditioning this process on historical observations, inferred SVO, and planned future trajectory of EV enables the synthesis of diverse, socially coherent future trajectories in a principled, multi-modal manner.

\section{METHODOLOGY}

\subsection{Problem Formulation}


We formulate trajectory prediction as the learning of a function that, given a short history of interactive motion, produces a socially coherent distribution over future trajectories.
Concretely, we draw $M$ samples \(\{\xi^{(m)},\mathbf{Y}^{(m)}\}_{m=1}^M\) from datasets, where each \(\xi^{(m)}\in\mathbb R^{F_h \times T_h}\) stacks the past $T_h$ timesteps of $F_h$ features, including longitudinal position \(x\), lateral position \(y\), and the  corresponding velocities \(v\) and acceleration $a$ for \(N\) vehicles in the scene.  The target future trajectory \(\mathbf{Y}^{(m)}\in\mathbb R^{F_f\times T_f}\) records the true positions $F_f=(x,y)$ over a horizon $T_f$.

\begin{figure*}[t]
  \centering
  \includegraphics[width=\linewidth]{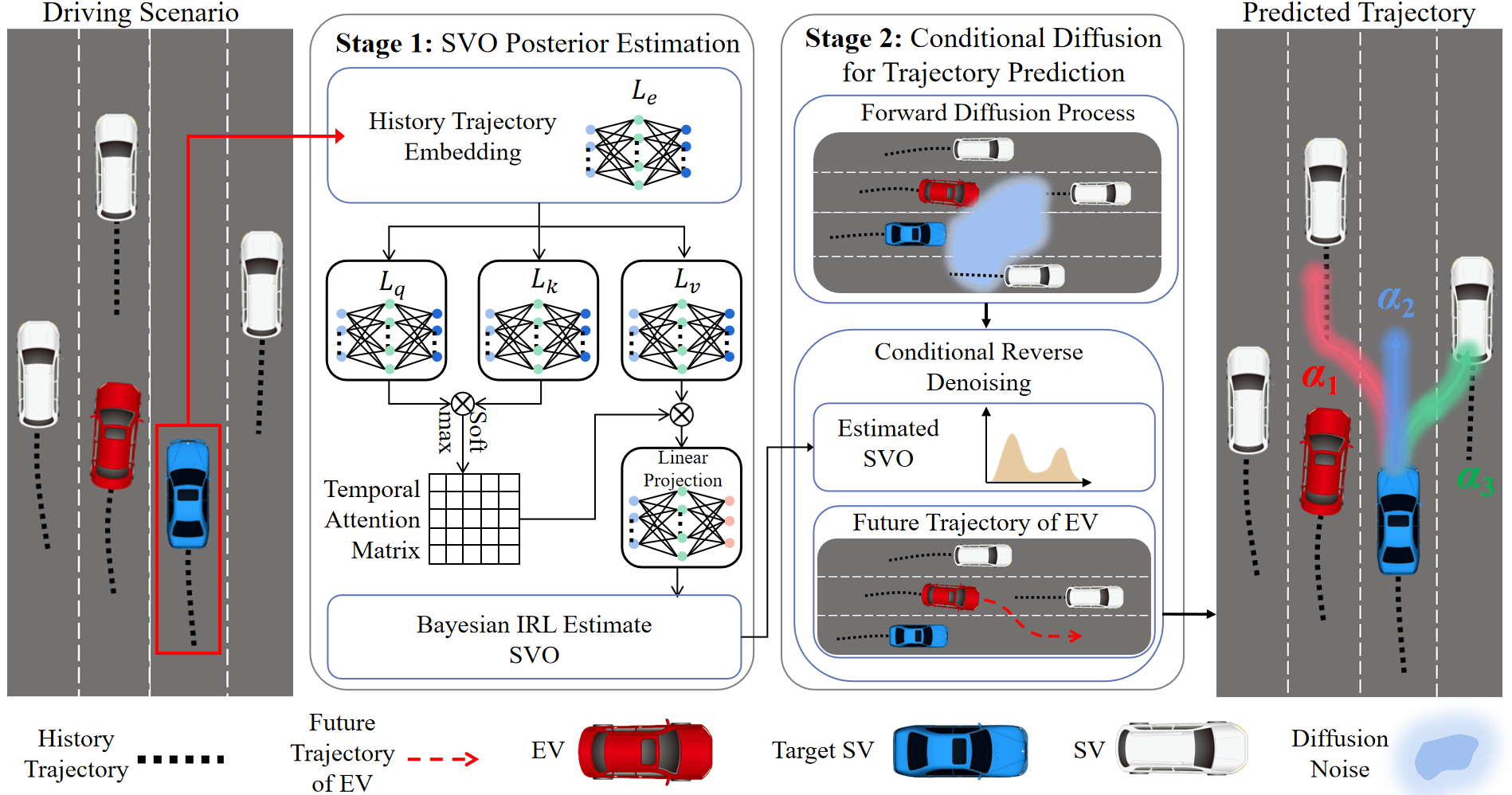}
  \caption{Overview of the SocialTraj framework.  Given the historical trajectories of the EV (red) and a target SV (blue) in a highway scenario, we first perform the attention‐based Bayesian IRL to infer the SVO of each SV.  The history trajectory is embedded by \(L_e\), transformed into queries, keys and values via \(L_q,L_k,L_v\), and aggregated through a temporal attention matrix to produce a context embedding, which is then converted into an SVO posterior via a linear projection and Bayesian IRL.  In the second stage, the conditional Diffusion module uses the forward diffusion process to corrupt ground truth trajectories, and then applies a conditional reverse denoising network guided by the estimated SVO and the planned future trajectory of the EV to generate multi-modal, socially coherent trajectory samples.}
  \label{fig:socialtraj_framework}
\end{figure*}

Our objective is to capture the intrinsic multi-modality of human driving in trajectory prediction.  {To this end, we decompose the prediction task into two stages as demonstrated in Fig. \ref{fig:socialtraj_framework}.} First, we infer latent SVOs of SVs via attention-based Bayesian IRL, thereby quantifying their cooperative or selfish tendencies. Second, we embed these SVO estimations, together with the planned trajectory of EV, into a conditional denoising diffusion model that aligns generated samples with historical driving styles and social context.  This two‐stage pipeline enables the learned prediction function to produce accurate, socially compliant trajectory distributions that adapt to both dynamic interactions and long‐term planning cues.  

\subsection{{Stage 1: SVO Posterior Estimation}}

Accurately inferring the social intent of each SV is essential to generate socially coherent trajectory predictions.  To this end, we first encode the recent motion history of SV into a rich, temporally attention-based representation and then perform Bayesian IRL to recover its SVO angle \(\alpha\in[0,2 \pi)\).

Concretely, given a history window \(\tau_H=\{(x_t,y_t,v_t,a_t)\}_{t=1}^{H}\) of length \(H=T_h\), we construct a sequence of four‐dimensional feature vectors \(\tau_t=[x_t,y_t,v_t,a_t]^\top\).  A shared embedding layer \(L_e\colon\mathbb R^4\to\mathbb R^{d_x}\) projects each \(\tau_t\) into a \(d_x\)–dimensional latent vector \(E_t=L_e(\tau_t)\), where $d_x$ denotes the dimension of the feature embedding. The embedded latent vectors are then stacked in \(E\in\mathbb R^{H\times d_x}\).

To discern which timesteps most strongly reflect cooperative or competitive behavior, we employ a temporal self‐attention module.  Specifically, three learnable projections \(L_q,L_k\colon\mathbb R^{d_x}\to\mathbb R^{d_k}\) and \(L_v\colon\mathbb R^{d_x}\to\mathbb R^{d_v}\) transform \(E\) into
\[
Q = E\,W_q,\quad K = E\,W_k,\quad V = E\,W_v,
\]
where \(W_q,W_k\in\mathbb R^{d_x\times d_k}\) and \(W_v\in\mathbb R^{d_x\times d_v}\).  The unnormalized attention scores \(S=QK^\top\in\mathbb R^{H\times H}\) require \(O(H^2d_k)\) operations.  Scaling by \(1/\sqrt{d_k}\) and normalizing each row with softmax, the attention matrix \(A\) is defined as:
\begin{equation}
A_{t,j}=\frac{\exp\bigl(S_{t,j}/\sqrt{d_k}\bigr)}{\sum_{j'=1}^H\exp\bigl(S_{t,j'}/\sqrt{d_k}\bigr)}
\end{equation}
The resulting context matrix
\(
C = A\,V
\)
encodes, for each timestep \(t\), a \(d_v\)–dimensional summary \(c_t\) that aggregates information from all history steps weighted by their relevance to the latent intent of the SV.

Building on this attention-based representation, we parameterize the instantaneous reward of SV $i$ as a weighted sum of egoistic and social objectives as:
\begin{equation}
R_\alpha^i(c_t)
=\cos\alpha\,R_{\mathrm{i}}^i(c_t)
+\sin\alpha\,R_{\mathrm{g}}^i(c_t)
\end{equation}
where \(R_{\mathrm{i}}^i\) might penalize deviation from a desired speed or large accelerations, while \(R_{\mathrm{g}}^i\) rewards safe headways and considerate lane changes.  This trigonometric form ensures a smooth interpolation between purely selfish (\(\alpha=0\)) and purely altruistic (\(\alpha=\tfrac\pi2\)) behaviors.

Under the maximum‐entropy IRL framework, the likelihood of observing \(\tau_H\) under SVO angle \(\alpha\) becomes
\begin{equation}
P(\tau_H\mid\alpha)
=\frac{1}{Z(\alpha)}\exp\Bigl(\sum_{t=o}^{H-1}\gamma^{\,t-1}R_\alpha(c_t)\Bigr)
\end{equation}
where \(\gamma\in(0,1)\) discounts future rewards and \(Z(\alpha)\) normalizes over the space of all possible trajectories.  This formulation balances fidelity to the demonstrated actions with an entropy prior, thereby remaining robust to stochastic deviations in human driving.

Since driving contexts are highly dynamic, a single static estimation of \(\alpha\) is insufficient.  {We therefore maintain a posterior distribution \(p_k(\alpha)\) that evolves as new observations arrive in the $k$-th Bayesian update.}  Starting from a truncated Gaussian prior \(p_0(\alpha)=\mathcal T\mathcal N_{[0,\frac\pi2]}(\mu_0,\sigma_0^2)\), each new history segment \(\tau_H^{(k)}\) triggers a Bayesian update as:
\begin{equation}
p_k(\alpha)\;\propto\;p_{k-1}(\alpha)\;P\bigl(\tau_H^{(k)}\mid\alpha\bigr)
\end{equation}
This recursion enforces temporal consistency while allowing rapid adaptation to abrupt behavioral shifts, such as a sudden yield or acceleration, within a single observation window.

In practice, we approximate the partition function \(Z(\alpha)\) via a Laplace expansion around the most likely trajectory, and draw samples \(\{\alpha^{(n)}\}\) from the posterior using Hamiltonian Monte Carlo (HMC) with adaptive leapfrog steps to ensure efficient exploration.  The posterior mean \(\hat\alpha=\mathbb E_{p_k}[\alpha]\) serves as our SVO estimation, and the posterior variance quantifies epistemic uncertainty.

By integrating attention-driven context embeddings with Bayesian IRL, our method delivers interpretable SVO estimations that are both robust to observation noise and responsive to evolving social dynamics.  The resulting posterior over \(\alpha\) feeds directly into our conditional diffusion model, ensuring that generated trajectories faithfully reflect the latent social preference of each driver and the confidence therein. 

\subsection{Stage 2: Conditional Diffusion for Trajectory Prediction}

Leveraging the SVO estimations derived from our attention-based Bayesian IRL framework, we employ a conditional DDPM to generate socially coherent and multi-modal trajectory predictions. Let $x_0 \in \mathbb{R}^{T_f \times 2}$ represent the ground-truth future trajectory of a target SV over $T_f=25$ timesteps in the $(x,y)$ plane. Our approach integrates forward diffusion, conditional reverse denoising, and uncertainty decomposition, resulting in interactive and socially-aware predictions.

During the forward diffusion process, the trajectory $x_0$ is progressively corrupted through a discrete Gaussian diffusion over $T_s$ steps as follows:
\begin{equation}
q(x_{1:T_s}\mid x_0) = \prod_{t=1}^{T_s}\mathcal{N}\left(x_t; \sqrt{1-\phi_t}\, x_{t-1}, \phi_t I_2\right)
\end{equation}
where the noise schedule $\{\phi_t\}_{t=1}^{T_s}$ adheres to the cosine variance approach proposed by \cite{nichol2021improved}, improving training stability. As $t \to T_s$, the corrupted state $x_t$ converges toward an isotropic Gaussian distribution $\mathcal{N}(0, I_2)$.

For trajectory reconstruction, the reverse denoising employs a neural denoiser $\epsilon_\phi$ explicitly conditioned on the feature set $\psi$, encompassing historical observations, the estimated SVO angle $\alpha$, and the planned future trajectory of the EV. Specifically, given a noisy trajectory state $x_t$, the conditional reverse step as follows:
\begin{equation}
x_{t-1} = \frac{1}{\sqrt{\gamma_t}}\left(x_t - \frac{1-\gamma_t}{\sqrt{1-\bar\gamma_t}}\epsilon_\phi(x_t, t, \psi)\right) + \sqrt{\delta_t}\, \zeta_t
\end{equation}
where $\gamma_t = \bar\gamma_t/\bar\gamma_{t-1}$, $\delta_t$ denoting the noise variance at timestep $t$, $\bar\gamma_t = \prod_{s=1}^t(1 - \delta_s)$, and $\zeta_t \sim \mathcal{N}(0, I_2)$. By integrating historical trajectory information, inferred social preferences $\alpha$, and the planned trajectories of EV into the conditioning set $\psi$, the denoiser effectively learns contextually appropriate interactions among vehicles. This approach ensures that higher SVO values (altruistic behaviors) promote cooperative maneuvers, while lower SVO values (egoistic behaviors) encourage assertive driving styles.

The denoiser $\epsilon_\phi$ is trained by minimizing a noise-prediction loss function defined as:
\begin{equation}
\mathcal{L}(\phi) = \mathbb{E}_{t,x_0,\epsilon,\psi}\left[\|\epsilon - \epsilon_\phi(\tilde{x}_t, t, \psi)\|^2\right]
\end{equation}
where $\tilde{x}_t = \sqrt{\bar\gamma_t}x_0 + \sqrt{1 - \bar\gamma_t}\epsilon$, and the expectation includes uniformly sampled diffusion timesteps, ground-truth trajectories, noise, and conditional features $\psi$. Incorporating estimated SVO within $\psi$ directly embeds epistemic uncertainty about social intent into the denoiser, facilitating predictions that reflect diverse driving behaviors without explicit retraining.

At inference, we sample multiple SVO estimations $\{\alpha^{(k)}\}\sim P(\alpha\mid\tau_H)$ through HMC, capturing epistemic uncertainty about latent social intentions. Each SVO sample yields distinct trajectory predictions through reverse diffusion initiated from Gaussian noise. Consequently, the ensemble $\{\hat{x}_0^{(k)}\}$ inherently captures two uncertainty types: aleatoric uncertainty, resulting from the stochasticity inherent to the diffusion process, and epistemic uncertainty, arising from variability in the inferred SVO. Formally, predictive covariance at each timestep decomposes as:

\begin{equation}
\underbrace{\mathbb{E}_{\alpha}[\mathrm{Cov}(\hat{x}_0\mid\alpha)]}_{\text{Aleatoric}} + \underbrace{\mathrm{Cov}_{\alpha}[\mathbb{E}(\hat{x}_0\mid\alpha)]}_{\text{Epistemic}}    
\end{equation}

The HMC sampling effectively explores the posterior SVO distribution, enabling the trajectory generation to reflect diverse potential social behaviors comprehensively. This explicit uncertainty modeling supports risk-aware planning decisions in downstream autonomous driving applications, clearly distinguishing intrinsic environmental variability from social behavioral uncertainty.


{Through explicit conditioning on $\psi$, which integrates estimated social preferences and EV future trajectories, our conditional diffusion approach addresses common trajectory prediction challenges such as mode collapse and unrealistic behaviors. Coupling Bayesian IRL-driven social inference with conditional diffusion ensures generated trajectories are behaviorally plausible, socially consistent, and computationally efficient, thus providing a two-stage trajectory prediction framework for socially-aware autonomous driving.}

\subsection{Joint Training}

The joint training process harmonizes attention-based Bayesian IRL with conditional diffusion models, enabling socially aware and behaviorally coherent trajectory prediction. Initially, the Bayesian IRL module iteratively refines the posterior distribution of the SVO angle $\alpha$ using sequential historical trajectory windows $\tau_H^{(i)}$. At each iteration, the posterior $P(\alpha|\tau_H)$ is updated via HMC, effectively exploring the posterior space by dynamically adjusting proposal distributions. The angular constraint $\alpha \in [0, 2\pi)$ ensures interpretability and physical plausibility of the estimated SVO.

Concurrently, the conditional diffusion model is trained to denoise trajectory samples $x_t^{(k)}$ from a corrupted state to clear predictions $\hat{x}_0^{(k)}$, leveraging a conditioning feature set $\psi^{(k)}$ that integrates estimated SVO angles $\alpha^{(k)}$ and the planned future trajectories of the EV. Specifically, the SVO angles sampled from the posterior distribution $\alpha^{(k)} \sim P(\alpha|\tau_H)$ are encoded into embeddings $e^{(k)} = \mathcal{E}(\alpha^{(k)})$, forming a critical component of the conditioning feature set $\psi^{(k)}$.

Crucially, the SVO embeddings $e^{(k)}$ infuse social behavioral information directly into the diffusion process. This embedding also regularizes the IRL component, maintaining physically interpretable distributions for $\alpha$. The resulting bidirectional coupling ensures that uncertainty in social preferences propagates effectively into trajectory diversity, yielding predictions that faithfully capture both individual driving styles and collective social norms.

To enhance training stability and align trajectory prediction granularity with the increasing certainty in SVO estimations, we implement a curriculum scheduling strategy. Specifically, the number of diffusion steps $T$ gradually increases from 100 to 200 as the posterior variance of the SVO estimation decreases. This progressive complexity aligns the predictive capability of the model with the confidence in its social inference.
Algorithm \ref{alg:full_algorithm} provides a detailed overview of the proposed SocialTraj framework, encompassing the two-stage joint training process.

\begin{algorithm}[h]
\SetAlgoLined
\KwIn{Historical trajectory $\tau_H = \{s_1,a_1,\dots,s_H\}$, \\
\hspace*{2em} Diffusion steps $T$, \\
\hspace*{2em} Sample count $K$}
\KwOut{Predicted trajectories $X_{\text{pred}} = \{\hat{x}_0^{(k)}\}_{k=1}^K$}

\textbf{Stage 1: SVO Posterior Estimation}
\begin{enumerate}
\item Initialize prior distribution: \\
$\alpha \sim \mathcal{TN}_{[0,2 \pi)}(\mu_0, \sigma_0^2)$

\item \For{each historical window $\tau_H^{(i)} \subset \tau_H$}{
    Compute likelihood: \\
    $P(\tau_H^{(i)}|\alpha) \propto \exp\left(\sum_{t=1}^H \gamma^{t-1}R(s_t,a_t;\alpha)\right)$

    Update posterior via HMC: \\
    $\alpha^{(i+1)} \sim P(\alpha|\tau_H^{(1:i)}) \propto P(\tau_H^{(i)}|\alpha)P(\alpha|\tau_H^{(1:i-1)})$
}
\item Obtain final posterior: $P(\alpha|\tau_H) = \mathcal{TN}(\mu_{\text{post}}, \sigma_{\text{post}}^2)$

\end{enumerate}

\vspace *{0.5em}
\textbf{Stage 2: Conditional Diffusion for Trajectory Generation}
\begin{enumerate}
\item Sample social preferences: $\{\alpha^{(k)}\}_{k=1}^K \sim P(\alpha|\tau_H)$

\item \For{each $\alpha^{(k)}$}{
    Encode SVO: $e^{(k)} = \mathcal{E}(\alpha^{(k)})$ \\
    Initialize noise: $x_T^{(k)} \sim \mathcal{N}(0, \mathbf{I})$

    \For{$t = T$ \KwTo $1$}{
        Compute denoising step with conditioning set $\psi^{(k)}$: \\
        $\epsilon_t = \epsilon_\psi\left(x_t^{(k)}, t, \psi^{(k)}\right)$

        Compute drift term using noise schedule parameters \(\gamma_t\) and \(\delta_t\): \\
        $\mu_t = \frac{1}{\sqrt{\gamma_t}}\left(x_t^{(k)} - \frac{1-\gamma_t}{\sqrt{1-\bar{\gamma}_t}}\epsilon_t\right)$

        Update sample: \\
        $x_{t-1}^{(k)} \sim \mathcal{N}(\mu_t, \delta_t\mathbf{I})$
    }
    Store final prediction: $\hat{x}_0^{(k)} = x_0^{(k)}$
}

\end{enumerate}

\Return $X_{\text{pred}}$
\caption{{SocialTraj for Trajectory Prediction }}
\label{alg:full_algorithm}
\end{algorithm}

\section{EXPERIMENTS}
\subsection{Datasets}

\begin{figure}[htbp]
\centering
\subfigure[US101]{
\begin{minipage}[t]{0.3\linewidth}
\includegraphics[height=2.5cm]{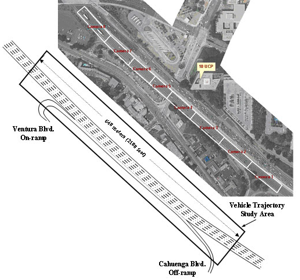}
\end{minipage}%
}%
\subfigure[I-80]{
\begin{minipage}[t]{0.35\linewidth}
\centering
\includegraphics[height=2.5cm]{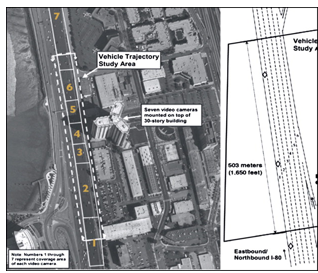}
\end{minipage}%
}%
\subfigure[HighD]{
\begin{minipage}[t]{0.3\linewidth}
\centering
\includegraphics[height=2.5cm]{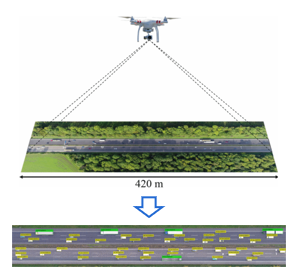}
\end{minipage}%
}%
\caption{Visualization of the three scenarios from the NGSIM and HighD datasets that are used in our experiments. The NGSIM dataset provides detailed aerial imagery of US101 (a) and I-80 (b) with overlaid vehicle trajectories. The HighD dataset offers drone‐captured footage of a 420\,m highway stretch (b) with the position and size of each vehicle annotated. Both datasets cover extensive, application‐oriented traffic scenarios.} 
\label{scenrios}
\end{figure}

We conduct experiments on the publicly available highway datasets Next Generation Simulation (NGSIM) \cite{punzo2011assessment} and Highway Drone (HighD) \cite{krajewski2018highd}, {both of which feature highly dynamic and interactive highway traffic scenarios as shown in Fig. \ref{scenrios}.}

\textbf{NGSIM}: This dataset consists of vehicle trajectory datasets
from US-101 and I-80, containing approximately 45 minutes of vehicle trajectory data at 10 Hz. It is critical for the analysis of vehicle behavior in a variety of traffic scenarios and assists in the development of reliable autonomous driving models.

\textbf{HighD}: It is a dataset of vehicle trajectories collected from six locations
on German highways. It includes 110,000 vehicles, including
cars and trucks, and a total distance traveled of 45,000 km.
This dataset provides detailed information about each vehicle,
including type, size, and maneuvers, making it invaluable for
advanced vehicle trajectory analysis and autonomous driving research.

Given the different sampling rates and data patterns inherent to the NGSIM and HighD datasets, we first standardize them by down‐sampling all trajectories to a uniform 5 Hz.  Each vehicle trajectory is then partitioned into 8 s clips, yielding 40 discrete time steps per scene with 15 history frames and 25 prediction frames.  In every clip, the initial 3 s of motion (15 samples) serve as the input history trajectory for the model, while the subsequent 5 s (25 samples) constitute the ground‐truth future trajectory.  Finally, to facilitate robust model evaluation, the processed dataset is split into training, validation, and test subsets according to a {7:1:2} ratio.  This data pre-processing ensures consistent temporal resolution and segment length across both NGSIM and HighD, enabling fair comparison of forecasting performance.


\subsection{Experimental Setup}
The experiments with the aforementioned datasets are conducted on the Windows 11 system environment with an Intel Core i7-14700KF processor and an NVIDIA RTX 4070 Ti GPU. Our model is developed using PyTorch and trained to converge based on the hyperparameters listed in Table \ref{tab:hyperparams_updated}.

\begin{table}[t]
\centering
\caption{Hyperparameters for the SocialTraj Framework}
\label{tab:hyperparams_updated}
\begin{tabular}{lll}
\toprule
\multicolumn{3}{c}{\textbf{Model I/O Settings}} \\ 
\midrule
History length & \(T_{\rm h}=15\) & Input length \\
Input dim.     & \(d_{\rm in}=4\) & \([x,y,v,a]\) \\
Pred. horizon  & \(T_{\rm f}=25\) & Output length \\
Output dim.    & \(d_{\rm out}=2\) & \([\hat{x},\hat{y}]\) \\
\midrule
\multicolumn{3}{c}{\textbf{Attention-based Bayesian IRL }} \\ 
\midrule
Embed dim.     & \(d_{\phi}=64\) & Encoder size \\
Attn. dim.     & \(d_q=d_k=d_v=64\) & Q/K/V dim. \\
HMC samples    & 200 & Posterior draws \\
Burn-in        & 200 & Discarded samples \\
Leapfrog steps & 10 & Per proposal \\
Step size      & \(\epsilon_0=1.0\) & Init. step size \\
Adapt rate     & \(\eta=0.01\) & Step decay \\
Acceptance     & 0.65 & Target rate \\
\midrule
\multicolumn{3}{c}{\textbf{SVO-conditional Diffusion}} \\ 
\midrule
Diffusion steps & 200 & Forward/reverse steps \\
Schedule offset & 0.008 & Cosine schedule \\
U-Net depth     & 5 & Blocks \\
Channels/block  & [64,128,256,512] & Feature maps \\
Kernel size     & 3 & Conv. kernel \\
Dilations       & [1,2,4,8] & Dilation rates \\
SVO embed dim.  & \(d_e=16\) & \(\mathcal{E}(\alpha)\) size \\
AdaLN dim.      & 64 & Cond. embed size \\
Batch size      & 64 & Per update \\
Optimizer       & AdamW & -- \\
Learning rate   & \(2\!\times\!10^{-4}\) & Init. LR \\
EMA decay       & 0.9999 & -- \\
\bottomrule
\end{tabular}
\end{table}

To evaluate the performance of the proposed SocialTraj framework on the NGSIM and HighD datasets, three widely adopted trajectory‐prediction metrics are employed as the evaluation metrics, including root mean square error (RMSE), average displacement error (ADE) and final displacement error (FDE).
RMSE is defined as $\sqrt{\frac{1}{T_f}\sum_{t=1}^{T_f}\bigl[(\hat{x}_t - x_t)^2 + (\hat{y}_t - y_t)^2\bigr]}\!$
where \(T_f\) is the prediction horizon, \((x_t,y_t)\) is the ground‐truth position at time \(t\), and \((\hat{x}_t,\hat{y}_t)\) is the prediction output of the model. RMSE captures overall stability by penalizing large per‐step deviations. Furthermore, ADE $= \frac{1}{T_f}\sum_{t=1}^{T_f}\bigl\lVert(\hat{x}_t,\hat{y}_t)-(x_t,y_t)\bigr\rVert_2$
measures the accuracy of the mean trajectory by averaging the Euclidean distance between the predicted and true trajectories in all time steps, while $\mathrm{FDE} = \bigl\lVert(\hat{x}_{T_f},\hat{y}_{T_f}) - (x_{T_f},y_{T_f})\bigr\rVert_2$ focuses specifically on the error of the endpoint in the final prediction step, thus evaluating the ability of the model to forecast the final position of the vehicle.
These metrics together provide a comprehensive assessment of predictive performance.

We compare and evaluate the proposed method against the following trajectory prediction models based on the evaluation metrics.
\begin{itemize}
    \item S-LSTM \cite{alahi2016social}: It uses an LSTM encoder-decoder for vehicle trajectory prediction, with a fully connected social pooling layer to predict future trajectories.
    \item CS-LSTM \cite{Deo2018Conv}: It builds on S-LSTM with a convolutional social pooling layer to capture interactions and incorporates multi-modal trajectory prediction based on horizontal and vertical driving intentions.

    \item S-GAN \cite{gupta2018social}: It combines sequence prediction with GANs, generating multiple trajectory predictions and using the closest to the true future trajectory for evaluation.
    \item WSiP \cite{Wang2023Wsip}: Inspired by wave superposition, this model aggregates local and global vehicle interactions for dynamic social pooling.

    \item PiP \cite{Song2020pip}: It considers the impact of EV planning on nearby vehicle trajectories using an LSTM encoder and a convolutional social pooling module.

    \item S-TF \cite{Liu2023stf}: It utilizes a sparse Transformer for multi-modal prediction, incorporating trajectory, velocity, and acceleration information along with driving intentions (left offset, right offset, or straight).

\end{itemize}
Table \ref{tab:baseline_comparison} summarizes the differences in modules and architectures between our model and baseline models.

\begin{table*}[t]
\centering
\caption{Comparison of Baseline Models and Ours IN MODULES AND ARCHITECTURES}
\label{tab:baseline_comparison}
\begin{tabular}{lcccccc}
\toprule
\textbf{Model}   & \textbf{Motion} & \textbf{Interaction} & \textbf{Multi-modal} & \textbf{Ego‐Plan} & \textbf{Generative} & \textbf{Enc–Dec} \\
                 & \textbf{Feature} & \textbf{Feature}     & \textbf{Output}     & \textbf{Condition} & \textbf{Framework}  & \textbf{Backbone} \\ 
\midrule
S‐LSTM           & \(\checkmark\) &            &    &   &    & \(\checkmark\) \\
CS‐LSTM          & \(\checkmark\) & \(\checkmark\)       & \(\checkmark\)   &   &    & \(\checkmark\) \\
S‐GAN            & \(\checkmark\) &            & \(\checkmark\)   &   & GAN           & \(\checkmark\) \\
WSiP             & \(\checkmark\) & \(\checkmark\)       & \(\checkmark\)   &   &    & \(\checkmark\) \\
PiP              & \(\checkmark\) & \(\checkmark\)       & \(\checkmark\)   & \(\checkmark\)  &    & \(\checkmark\) \\
S‐TF             & \(\checkmark\) & \(\checkmark\)       & \(\checkmark\)   & \(\checkmark\)  &    & \(\checkmark\) \\
\textbf{Ours}    & \(\checkmark\) & \(\checkmark\)       & \(\checkmark\)   & \(\checkmark\)  & Diffusion     & \(\checkmark\) \\
\bottomrule
\end{tabular}
\end{table*}

\subsection{Case Study of SVO estimation}

SVO provides a principled framework for the EV to infer the social preferences of SVs, thereby enabling socially coherent trajectory predictions.
In our framework, we leverage an attention‐based Bayesian IRL algorithm to infer and update the belief of EV over the SVO of SV from the history trajectory.  
To verify the effectiveness of SVO in capturing the multi-modality of driving styles in interactive traffic, we analyze a merge scenario drawn from the NGSIM I-80 dataset. 

\begin{figure}[t]
  \centering
  \includegraphics[width=\linewidth]{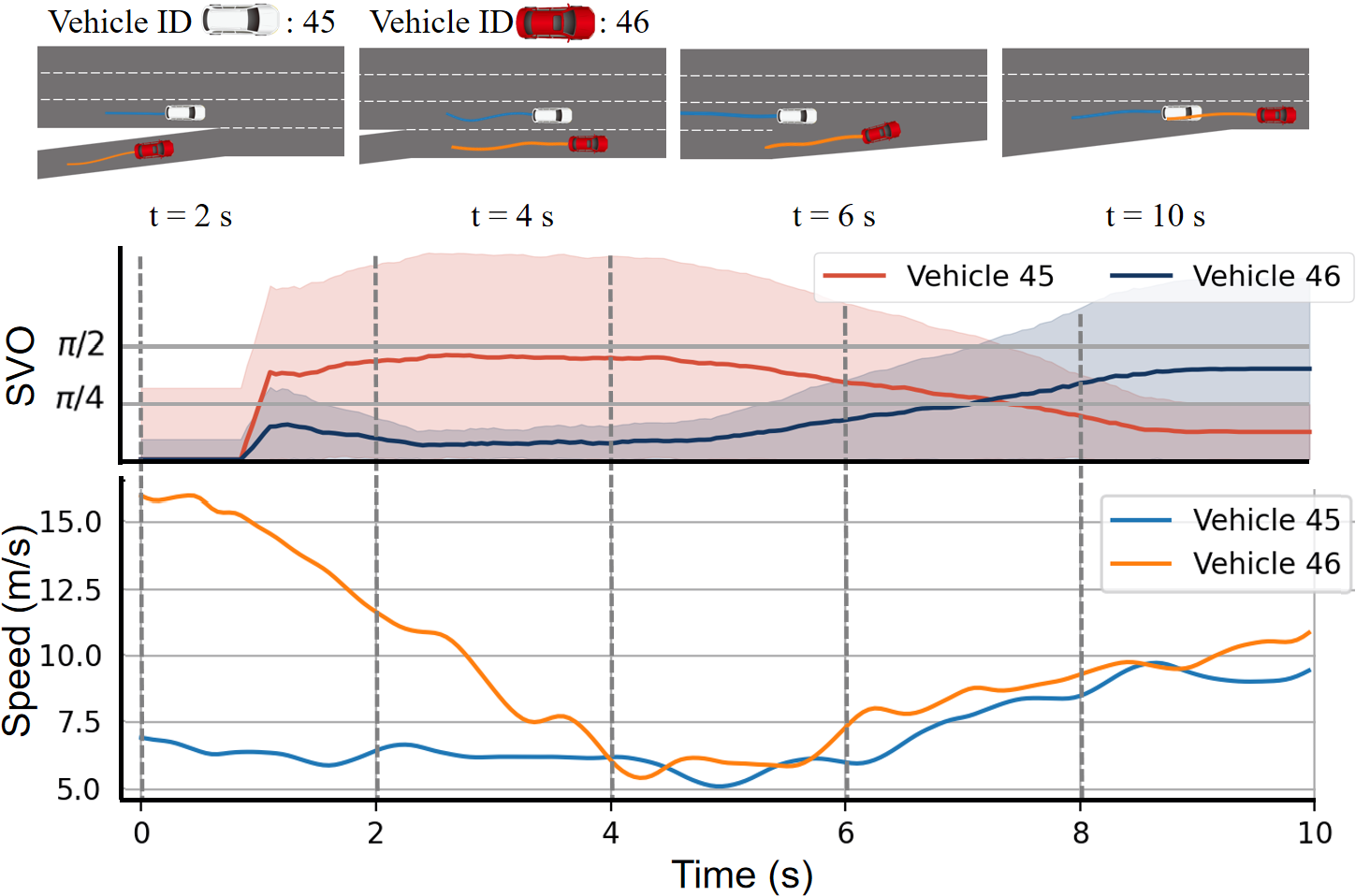}
  \caption{%
    (Top) Snapshot of NGSIM I-80 dataset with two interactive vehicles, where the EV (vehicle ID 46) seeks to merge into a lane occupied by the target SV (vehicle ID 45). 
    (Middle) For SVO estimations at each frame, the solid lines represent the inferred SVO, and the shaded region represents the confidence bounds.  
    (Bottom) Corresponding speed profiles of EV and SV, illustrating the yielding deceleration (\(t\approx2\text{–}6\) s) and re-acceleration of EV, alongside the gradual speed increase of SV.
  }
  \label{fig:merge_svo_speed}
\end{figure}

{In the illustrative merge scenario shown in Fig. \ref{fig:merge_svo_speed}, the EV (vehicle ID 46) initiates a lane change into the adjacent lane occupied by the SV (vehicle ID 45). At the beginning, the EV cruises at approximately 16 m/s, while the SV maintains a steady speed of around 7 m/s. Correspondingly, the attention-based Bayesian IRL algorithm initially assigns low SVO angles ($\theta \approx 0$) to both vehicles, reflecting purely self-interested driving styles and avoiding risky decisions due to optimistic estimations.
By $t \approx 1$ s, the EV begins a pronounced deceleration, which represents a partial shift toward cooperative behavior. Quantitatively, the posterior mean SVO of EV rises to just under $\pi/4$. Nevertheless, its speed remains considerably higher than that of the target SV, demonstrating that self-reward continues to dominate decision-making. In contrast, the SV remains at a low speed, and its inferred SVO rapidly ascends toward approximately $3\pi/4$, signifying strong altruistic behavior as it yields space to facilitate the EV's merging maneuver.
Once the EV completes its lane change at around $t \approx 6$ s, both vehicles accelerate. The EV regains speed to about $10$ m/s, while the SV also increases its speed, benefiting from the newly available clear road ahead. Reflecting these dynamics, the estimated SVO of the EV continues to rise toward $3\pi/4$, capturing the mutually beneficial aspect of its speed gain. Conversely, the SVO of the target SV decreases toward $\pi/8$, mirroring its renewed prioritization of self-reward.
This temporal coupling between velocity profiles and SVO posteriors demonstrates the power of our attention-based Bayesian IRL. By continually integrating the most salient past behaviors of each vehicle via temporal attention and updating social preferences in real time, SocialTraj faithfully tracks evolving driver intents and delivers socially coherent, multimodal trajectory predictions even in tightly interacting traffic.}



\subsection{Main Results}

We report the quantitative comparison with six baselines on the NGSIM and HighD datasets, using evaluation metrics including minADE, minFDE, and RMSE at horizons from 1 s to 5 s.  Bold entries indicate the best performance in each column. 
{As shown in Table \ref{tab:sota_comparison}, the proposed SocialTraj framework achieves a minADE of {0.28 m} and a minFDE of {0.71 m} on the NGSIM dataset, representing relative improvements of 20.0 \% and 19.3 \% over the strongest baseline S-TF (0.35 m/0.88 m), which records 0.35 m and 0.88 m, respectively.  Looking at the RMSE presented in Table \ref{tab:RSM_NGSIM}, our framework reduces the error of 1s from 0.99 m to 0.32 m, corresponding to a 67.7 \% reduction, and the 5 s error from 3.33 m to {2.85 m}, representing a 14.5 \% decrease.  These outcomes confirm that SocialTraj limits error growth more effectively over longer prediction horizons.}


{Similarly, SocialTraj achieves a minAD of 0.24 m, which represents a 22.6\% reduction relative to the 0.31 m recorded by S-TF on the HighD dataset as demonstrated in Table \ref{tab:sota_comparison}. It also achieves a minFDE of 0.62 m, corresponding to an improvement of 18.4\% over 0.76 m. The RMSE results reported in Table \ref{tab:RSM_HighD} echo this trend. In the first second, our error falls from 0.75 m to 0.11 m, giving an 85.3\% decrease, and in the fifth second it decreases from 1.75 m to 1.04 m, a 40.6\% reduction. These gains under both the high-density, high-speed conditions of the NGSIM dataset and the sparser freeway scenarios of the HighD dataset demonstrate the robustness and effectiveness of our approach.}

When compared to purely history‐based predictors like S-LSTM, CS-LSTM, S-GAN, and WSiP, the superior accuracy of our  SocialTraj framework highlights the vital role of social inference. By inferring the evolving SVO of each agent and conditioning trajectory generation on these beliefs, our model anticipates cooperative or adversarial maneuvers that would otherwise be missed.  Against PiP, which also incorporates ego‐planning cues, we further reduce minADE by 26.3 \% on NGSIM and 29.4 \% on HighD, underscoring the benefit of our {attention-based} Bayesian IRL over the static pooling of PiP.  Finally, relative to S-TF, which is the strongest single‐stage encoder, the SocialTraj framework lowers average RMSE by 67.7 \% at 1 s and 14.5 \% at 5 s on NGSIM, and by 85.3 \% and 40.6 \% on HighD, demonstrating that explicit, two‐stage social modeling with diffusion sampling yields substantial error reductions.

\begin{figure}[htbp]
    \centering
    \subfigure[US101]{
    \begin{minipage}[t]{\linewidth}
    \includegraphics[width=\linewidth]{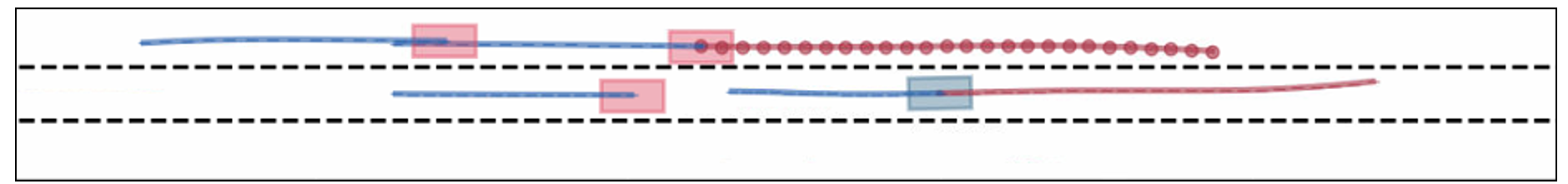}
    \end{minipage}%
    }\\
    \subfigure[I-80]{
    \begin{minipage}[t]{\linewidth}
    \centering
    \includegraphics[width=\linewidth]{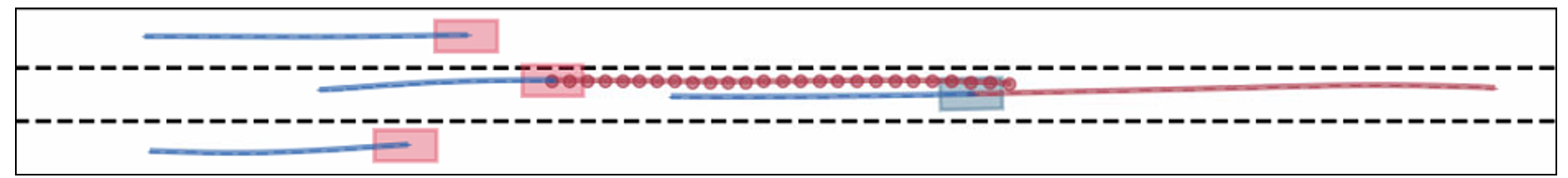}
    \end{minipage}%
    }\\
    \subfigure[HighD]{
    \begin{minipage}[t]{\linewidth}
    \centering
    \includegraphics[width=\linewidth]{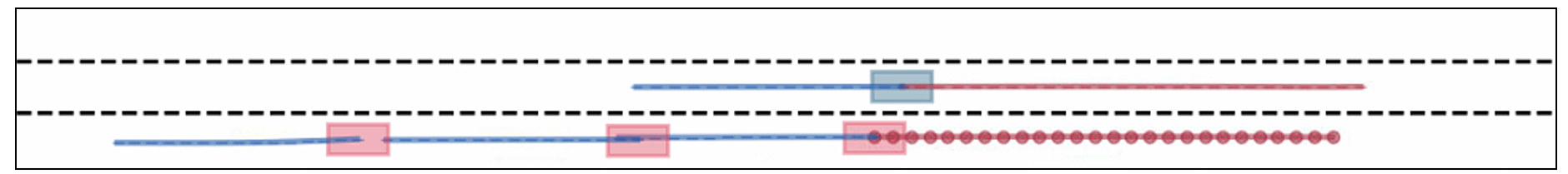}
    \end{minipage}%
    }%
  \caption{Qualitative prediction results in three representative driving scenarios.  Panels (a) and (b) show US‐101 and I‐80 driving interactions from the NGSIM dataset, while (c) depicts a highway segment from HighD.  In each subplot, we restrict the view to the two adjacent lanes of the EV (blue).  All vehicles display their observed histories over $T_h$ frames (solid blue lines), and the planned future trajectory of the EV is overlaid in a solid red line.  The predicted trajectory of target SV is drawn as dotted lines, with each dot marking a future timestep.  By conditioning on both the dynamically estimated SVO of the target SV and the ego-planning of EV, SocialTraj produces trajectories that closely track the ground truth across these diverse scenarios, demonstrating its ability to capture bidirectional social influences and generate socially coherent, multi-modal forecasts.}
  \label{fig:three_scenes}
\end{figure}

To further illustrate the performance of the SocialTraj framework, we visualize its highest-probability predictions in three scenarios on NGSIM and HighD, as shown in Fig. \ref{fig:three_scenes}. This demonstrates that SocialTraj generates trajectories that closely align with human driving, respecting both safety margins and lane boundaries. These improvements stem from three core innovations of SocialTraj: (i) {attention-based Bayesian IRL} provides real‐time, uncertainty‐aware estimations of the SVO of SVs; (ii) {SVO‐conditional diffusion} enables the sampler to focus on socially plausible trajectory modes rather than averaging over all possibilities; and (iii) the {denoising diffusion backbone} naturally captures multi-modality  and maintains predictive fidelity across both short‐term and long‐term horizons.  Together, these components deliver a unified framework that consistently outperforms baselines by significant margins in both accuracy and reliability.


\begin{table}[t]
\centering
\caption{Comparison with EACH MODEL on NGSIM and HighD dataset}
\label{tab:sota_comparison}
\begin{tabular}{lcccc}
\toprule
\textbf{Model}   & \multicolumn{2}{c}{\textbf{NGSIM}} & \multicolumn{2}{c}{\textbf{HighD}} \\
\cmidrule(lr){2-3}\cmidrule(lr){4-5}
                 & \textbf{minADE } & \textbf{minFDE} & \textbf{minADE} & \textbf{minFDE} \\
\midrule
S‐LSTM           & 0.54 & 1.21 & 0.48 & 1.05 \\
CS‐LSTM          & 0.47 & 1.10 & 0.42 & 0.98 \\
S‐GAN            & 0.43 & 1.02 & 0.39 & 0.91 \\
WSIP             & 0.41 & 0.98 & 0.37 & 0.87 \\
PiP              & 0.38 & 0.92 & 0.34 & 0.81 \\
S‐TF             & 0.35 & 0.88 & 0.31 & 0.76 \\
\textbf{Ours}    & \textbf{0.28} & \textbf{0.71} & \textbf{0.24} & \textbf{0.62} \\
\bottomrule
\end{tabular}
\end{table}

\begin{table}[ht]
  \centering
  \caption{RSME Comparison of Each Model on NGSIM Dataset}
  \label{tab:RSM_NGSIM}
  \resizebox{\columnwidth}{!}{%
    \begin{tabular}{@{}c|*{7}{c}@{}}
      \toprule
      \makecell[c]{Prediction\\duration}
        & S-LSTM & CS-LSTM & S-GAN & WSiP & PiP & S-TF & \textbf{Ours} \\
      \midrule
      1s & 0.60 & 0.58 & 0.57 & 0.56 & 0.55 & 0.99 & \textbf{0.32} \\
      2s & 1.28 & 1.26 & 1.32 & 1.23 & 1.18 & 1.43 & \textbf{0.83} \\
      3s & 2.09 & 2.07 & 2.22 & 2.05 & 1.94 & 1.70 & \textbf{1.17} \\
      4s & 3.10 & 3.09 & 3.26 & 3.08 & 2.88 & 2.02 & \textbf{1.76} \\
      5s & 4.37 & 4.37 & 4.40 & 4.34 & 4.04 & 3.33 & \textbf{2.85} \\
      \bottomrule
    \end{tabular}%
  }
\end{table}

\begin{table}[ht]
  \centering
  \caption{RSME Comparison of Each Model on HighD Dataset}
  \label{tab:RSM_HighD}
  \resizebox{\columnwidth}{!}{%
    \begin{tabular}{@{}c|*{7}{c}@{}}
      \toprule
      \makecell[c]{Prediction\\duration}
        & S-LSTM & CS-LSTM & S-GAN & WSiP & PiP & S-TF & \textbf{Ours} \\
      \midrule
      1s & 0.19 & 0.19 & 0.30 & 0.20 & 0.17 & 0.75 & \textbf{0.11} \\
      2s & 0.57 & 0.57 & 0.78 & 0.60 & 0.52 & 0.89 & \textbf{0.29} \\
      3s & 1.18 & 1.16 & 1.46 & 1.21 & 1.05 & 1.05 & \textbf{0.65} \\
      4s & 2.00 & 1.96 & 2.34 & 2.07 & 1.76 & 1.33 & \textbf{0.91} \\
      5s & 3.02 & 2.96 & 3.41 & 3.14 & 2.63 & 1.75 & \textbf{1.04} \\
      \bottomrule
    \end{tabular}%
  }
\end{table}

\begin{figure*}[!htbp]
  \centering
  \subfigure[Vanilla DDPM]{
    \includegraphics[width=0.49\linewidth]{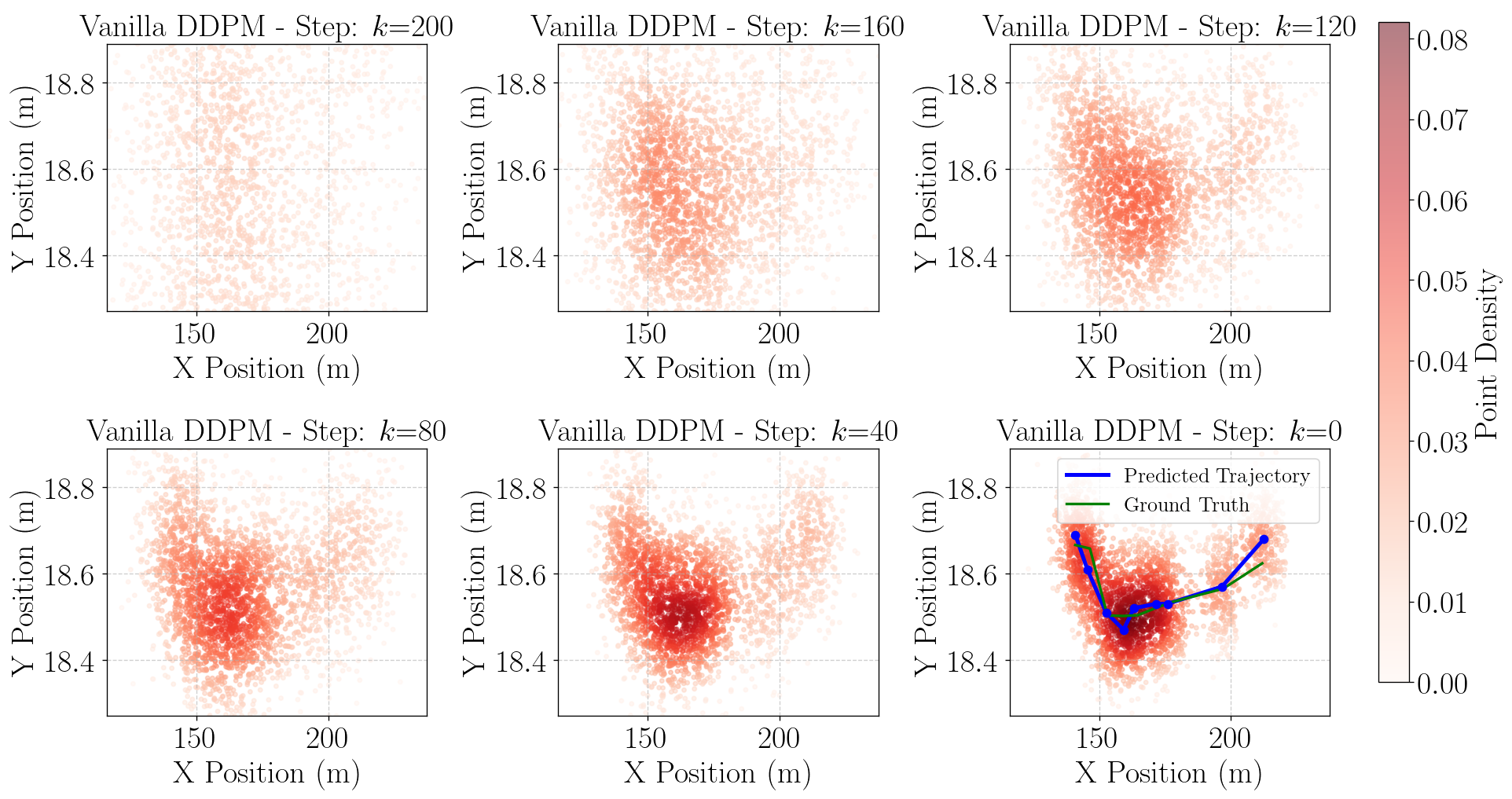}
    \label{fig:ablate_vanilla}
  }
  \hfill
  \subfigure[SVO-conditional DDPM ]{
    \includegraphics[width=0.46\linewidth]{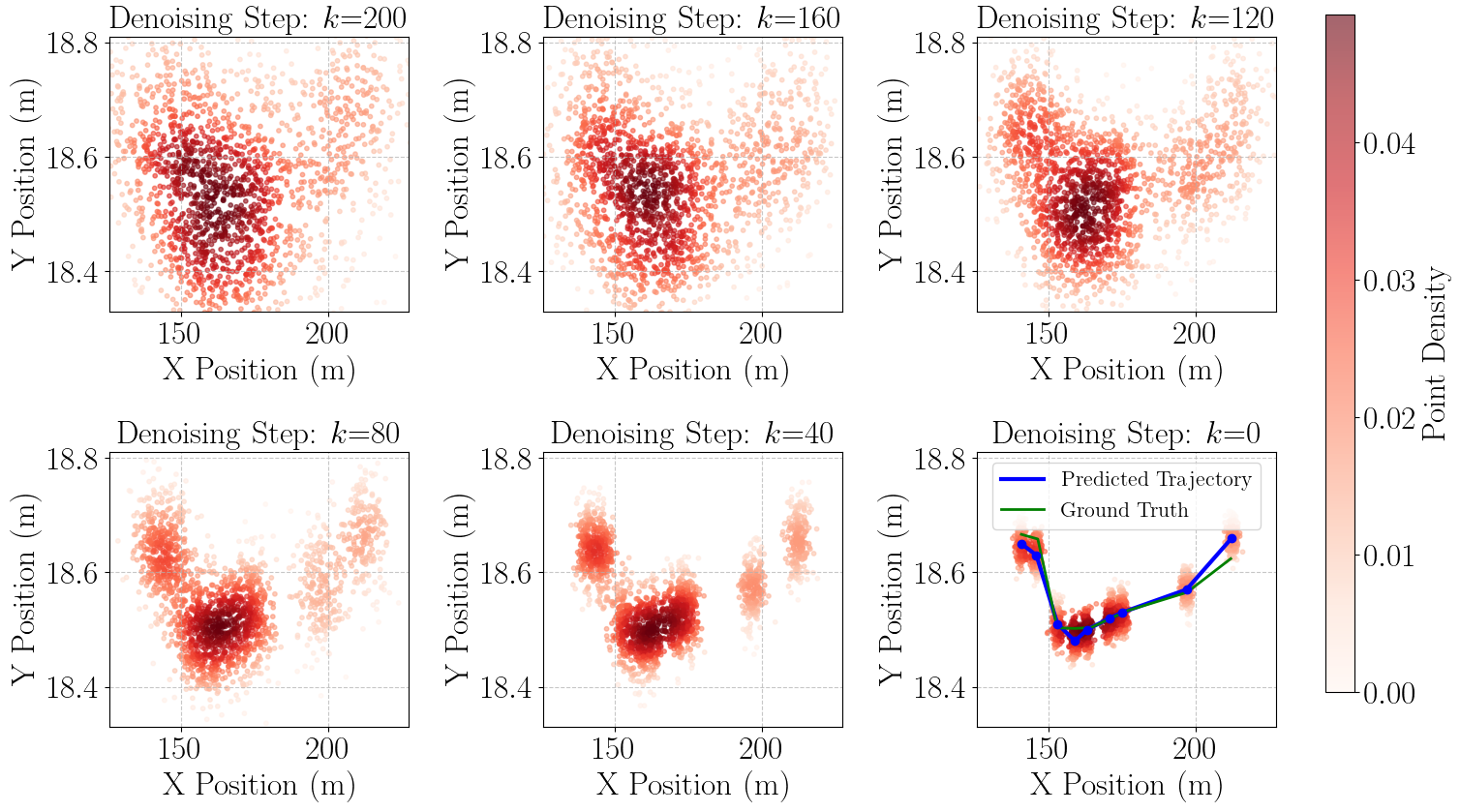}
    \label{fig:ablate_conditioned}
  }
  \caption{Comparison of denoising processes under ablation. Each subfigure shows six snapshots of sampled trajectories at diffusion steps \(k=\{200,160,120,80,40,0\}\). Point density is indicated by red shading, the blue line in the final subpanel traces the recovered trajectory, and the green line shows the ground truth trajectory. (a) The vanilla DDPM remains dispersed until the end, reflecting its lack of social or planning inputs. (b) Conditioning on estimated SVO and the planned trajectory of EV accelerates concentration into plausible lanes by \(k=120\) and yields accurate, multi-modal predictions by \(k=0\) using only 400 steps.}

  \label{fig:ablate_denoising}
\end{figure*}
\subsection{Ablation Study}

To dissect the individual contributions of the core components in our framework, including social inference, dynamic conditioning, and ego‐plan integration, we perform ablation studies that progressively remove or simplify each element.  
{These studies not only validate the necessity of each module but also reveal considerations in efficiency, adaptability, and predictive fidelity under the proposed SocialTraj framework.}

\subsubsection{DDPM Baseline}  
Replacing our conditional sampler with a vanilla DDPM (no social or ego-planning inputs) degrades performance to a minADE/minFDE of 0.41 m/0.98 m on NGSIM and 0.37 m/0.87 m on HighD as shown in Table~\ref{tab:ablation}.  This collapse toward mean predictions confirms that, without any social inference, the model cannot anticipate cooperative or adversarial maneuvers.  Moreover, we notice that our conditional sampler requires only 120 diffusion steps to achieve comparable generative quality while the vanilla DDPM requires 200 steps, resulting in a 40\% reduction in inference time while still preserving socially consistent and behaviorally aligned trajectory predictions.

By constraining the reverse diffusion to regions of high posterior probability that are informed by social context, the sampler converges more rapidly, reducing the number of Langevin or score‐matching iterations needed to denoise toward plausible trajectories.

Fig. \ref{fig:ablate_denoising} juxtaposesVisualization of the three scenarios from the NGSIM and HighD datasets that are used in our experiments. The NGSIM dataset provides detailed aerial imagery of US101 (a) and I-80 (b) with overlaid vehicle trajectories. The HighD dataset offers drone‐captured footage of a 420\,m highway stretch (b) with the position and size of each vehicle annotated. Both datasets cover extensive, application‐oriented traffic scenarios. the vanilla DDPM baseline against our SocialTraj framework. In Fig. \ref{fig:ablate_denoising} (a), the vanilla DDPM exhibits high dispersion until the final denoising stages, confirming its tendency to collapse toward average trajectories and its inability to leverage social context for faster convergence. Conversely, subfigure (b) demonstrates the benefits of SVO and EV-plan conditional diffusion model. 
At the initial step ($k=200$), samples are nearly isotropic around the prior mean, reflecting high aleatoric uncertainty. As the sampler progresses through $k=160$ and $k=120$, the cloud begins to elongate along the direction of the true lane geometry, illustrating that the model leverages SVO and EV-plan conditioning to prune unlikely directions. By $k=80$, three distinct mode clusters emerge, corresponding to the major lane options under the estimated social preferences. Further refinement at $k=40$ tightens each cluster and suppresses stray points. Finally, at $k=0$, the densest mode collapses into the solid blue trajectory, which matches the human-driven ground truth. This stark contrast highlights how social inference improves the predictive fidelity.

\subsubsection{Fixed SVO Conditioning}  
We next fix the SVO angle at three canonical values to represent distinct behavioral archetypes, thereby removing the dynamic adaptation of the attention-based Bayesian IRL module :
\begin{itemize}
  \item \emph{Pure egoism} (\(\alpha=0^\circ\)): Predictions yield minADE/minFDE of 0.33 m/0.80 m on NGSIM and 0.30 m/0.81 m on HighD.  The model underestimates yielding behavior, leading to overly aggressive trajectory prediction.
  \item \emph{Balanced cooperation} (\(\alpha=45^\circ\)): Error improves to 0.31 m/0.78 m and 0.28 m/0.76 m.  This setting best approximates human merging tendencies but still cannot adapt to on‐the‐fly changes.
  \item \emph{Pure altruism} (\(\alpha=90^\circ\)): Error worsens to 0.36 m/0.89 m and 0.32 m/0.82 m, as the model overestimates yielding and fails to exploit emerging gaps.
\end{itemize}
Although the \(45^\circ\) bias captures nominal cooperative maneuvers more accurately than the extremes, all fixed‐SVO variants remain 10–20\% worse than the proposed SocialTraj.  This gap highlights the importance of our attention‐based Bayesian IRL. Only by updating beliefs in real time can the model track rapid shifts in social intent,  which is essential in dense, highly interactive traffic.

\subsubsection{Without Future EV Plan}  
Finally, we omit the planned trajectory of EV from the conditioning set, relying solely on past histories and SVO.  Performance drops to 0.33 m/0.82 m on NGSIM and 0.29 m/0.79 m on HighD.  From an interaction standpoint, excluding the EV plan severs the explicit coupling: the sampler no longer knows whether the EV intends to yield or accelerate, so it cannot generate SV trajectories that anticipate or react to the forthcoming actions of EV.  By reinstating the EV plan, we reestablish bidirectional context, enabling socially coherent joint predictions and further reducing displacement errors.

\begin{table}[t]
  \centering
  \caption{Ablation Studies for Core Components}
  \label{tab:ablation}
  \begin{tabular}{@{}c
                  *{2}{>{\centering\arraybackslash}c}
                  *{2}{>{\centering\arraybackslash}c}
                  c@{}}
    \toprule
    \multirow{2}{*}{\textbf{Variant}}
      & \multicolumn{2}{c}{\textbf{NGSIM}}
      & \multicolumn{2}{c}{\textbf{HighD}}
      & \multirow{2}{*}{\textbf{Steps}} \\
      & ADE & FDE & ADE & FDE & \\
    \midrule
    DDPM (no social/plan)         & 0.41 & 0.98 & 0.37 & 0.87 & 200 \\
    Fixed \(\alpha=0^\circ\)      & 0.33 & 0.80 & 0.30 & 0.81 & 200 \\
    Fixed \(\alpha=45^\circ\)     & 0.31 & 0.78 & 0.28 & 0.76 & 200 \\
    Fixed \(\alpha=90^\circ\)     & 0.36 & 0.89 & 0.32 & 0.82 & 200 \\
    Without EV future plan        & 0.32 & 0.81 & 0.29 & 0.79 & 200 \\
    \textbf{SocialTraj (ours)}    & \textbf{0.37} & \textbf{0.95}
                                   & \textbf{0.32} & \textbf{0.86}
                                   & \textbf{120} \\
    \textbf{SocialTraj (ours)}    & \textbf{0.28} & \textbf{0.71}
                                   & \textbf{0.24} & \textbf{0.62}
                                   & \textbf{200} \\
    \bottomrule
  \end{tabular}
\end{table}

In summary, the ablations confirm that social inference and the conditional diffusion model are both necessary and complementary. Social context accelerates sampling and avoids mean collapse, fixed biases fail to adapt to evolving intent, and explicit EV plan inputs restore bidirectional interactive modeling.  Critically, the ability of our attention‐based Bayesian IRL to update SVO beliefs online underlies the superior performance of proposed SocialTraj framework in dynamic, complex driving scenarios.


\section{Conclusion}

In this work, we introduce SocialTraj, a socially aware, interactive two-stage trajectory prediction framework that brings together insights from social psychology and generative modeling to address the multi-modality and interactivity challenges of autonomous driving.  By inferring the SVO of each SV via an attention-driven Bayesian IRL module, we obtain a principled, uncertainty‐aware measure of cooperative intent.  Embedding these SVO estimations into a conditional denoising diffusion model enables modality‐aligned sampling of future trajectories that remain faithful to both historical driving style and social context.  Further, by incorporating the planned trajectory of EV, SocialTraj explicitly captures bidirectional influence, yielding richer, more coherent joint forecasts.  Extensive evaluations on the NGSIM and HighD benchmarks demonstrate that our approach not only outperforms a wide range of baselines across ADE, FDE, and RMSE metrics but also maintains real-time efficiency through reduced diffusion steps.  Together, these results validate the effectiveness of combining dynamic social inference with diffusion‐based generative sampling for accurate, socially compliant trajectory prediction in complex traffic environments. {Future works will include the extension of SocialTraj to more complex mixed traffic environments, including pedestrians and cyclists.}

\bibliographystyle{IEEEtran}
\bibliography{reference}

\begin{IEEEbiography}[{\includegraphics[width=1in,height=1.25in,clip,keepaspectratio]{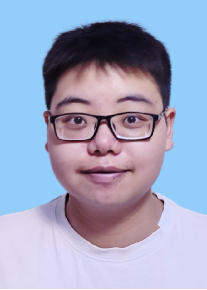}}]
{Xiao Zhou} received the B.Eng. degree in electronics and information engineering from Harbin Institute of Technology, Shenzhen, China, in 2023. She is currently pursuing the M.Phil. degree in robotics and autonomous systems at The Hong Kong University of Science and Technology, Guangzhou, China. Her research interests focus on deep reinforcement learning, optimization, decision-making, and motion planning, specifically applied to the fields of robotics and autonomous driving.
\end{IEEEbiography}

\begin{IEEEbiography}[{\includegraphics[width=1in,height=1.25in,clip,keepaspectratio]{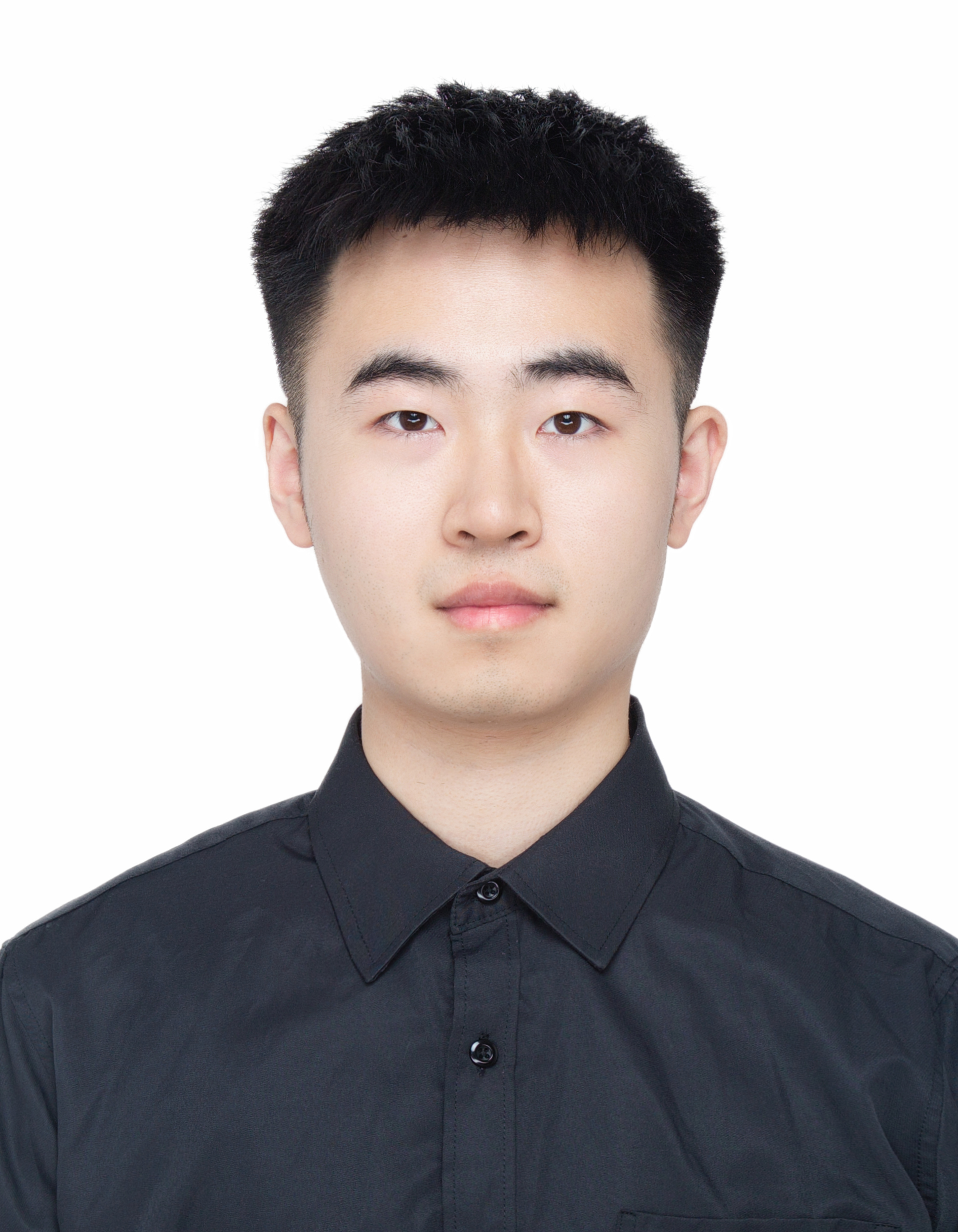}}]
{Zengqi Peng} received the B.Eng. degree in automation and the M.Sc. degree in artificial intelligence and automation from Huazhong University of Science and Technology, Wuhan, China, in 2019 and 2022, respectively. He is currently pursuing the Ph.D. degree in robotics and autonomous systems under the Guangzhou Pilot Scheme at The Hong Kong University of Science and Technology, China. His research interests include reinforcement learning, optimization, decision-making and motion planning with application to robotics and autonomous driving.
\end{IEEEbiography}

\begin{IEEEbiography}[{\includegraphics[width=1in,height=1.25in,clip,keepaspectratio]{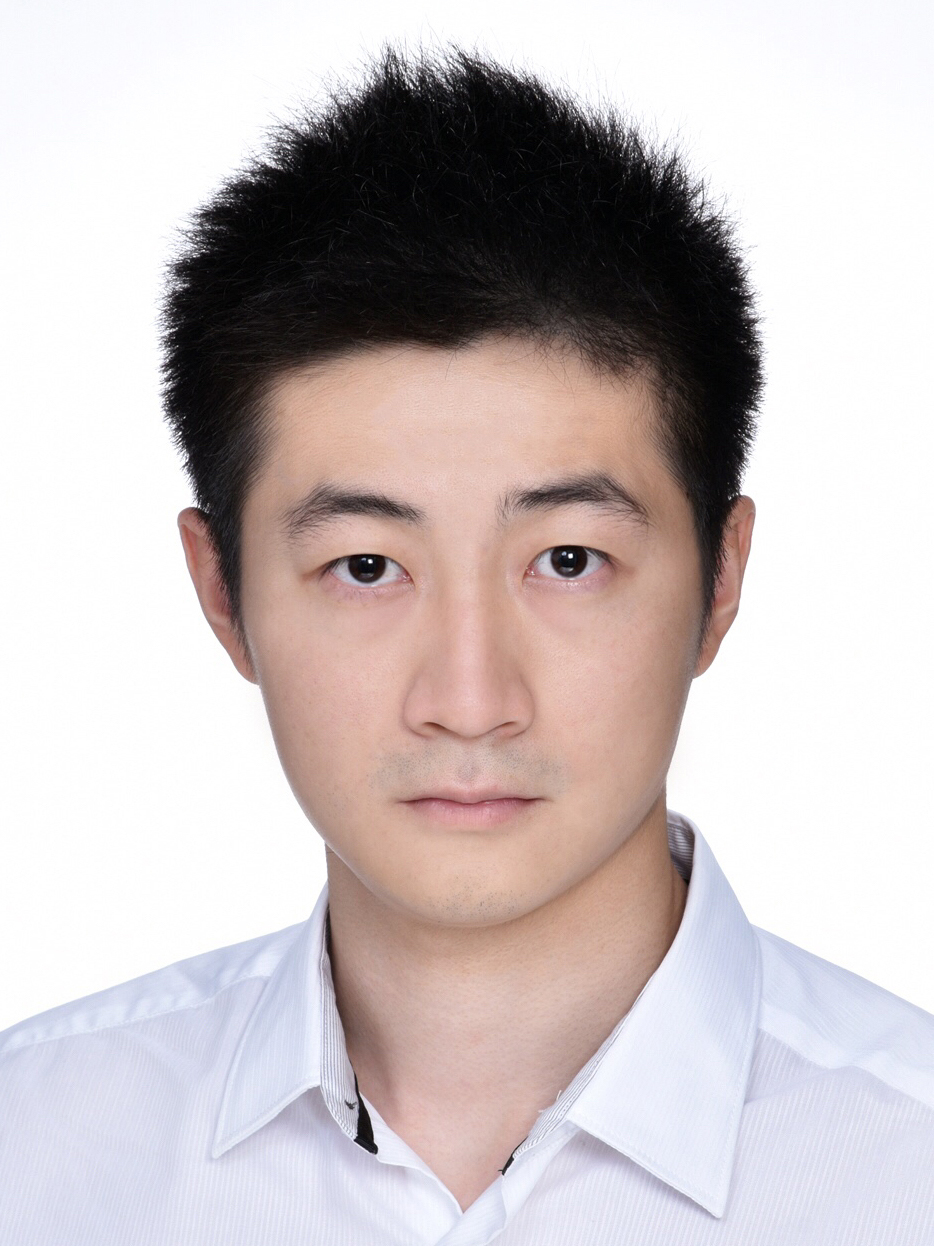}}]
	{Jun Ma} (Senior Member, IEEE) received the B.Eng. degree with First Class Honours in electrical and electronic engineering from Nanyang Technological University, Singapore, in 2014, and the Ph.D. degree in electrical and computer engineering from the National University of Singapore, Singapore, in 2018.
From 2018 to 2021, he held several positions at the National University of Singapore; University College London, London, U.K.; University of California, Berkeley, Berkeley, CA, USA; and Harvard University, Cambridge, MA, USA.  He is currently an Assistant Professor with the Robotics and Autonomous Systems Thrust, The Hong Kong University of Science and Technology (Guangzhou), Guangzhou, China, and also with the Division of Emerging Interdisciplinary Areas, The Hong Kong University of Science and Technology, Hong Kong SAR, China. He is also the Director of Intelligent Autonomous Driving Center, The Hong Kong University of Science and Technology (Guangzhou).
His research interests include motion planning and control for robotics and autonomous driving.
\end{IEEEbiography}

\end{document}